\documentclass{nature}
\usepackage{times}  %Required
\usepackage{helvet}  %Required
\usepackage{courier}  %Required
\usepackage{url}  %Required
\usepackage{epsfig}
\usepackage{graphicx}
\usepackage[fleqn]{amsmath}
\usepackage{algorithm, algorithmic}
\usepackage{amssymb}
\usepackage{amsmath}
\usepackage{color}

\title{EasiCSDeep: A deep learning model \\for Cervical Spondylosis Identification\\using
surface electromyography signal}
\author{Nana Wang$^{1,2}$, Li Cui$^{1,4}$, Xi Huang$^{1}$, Yingcong Xiang$^{1,2}$, Jing Xiao$^{3}$}

\begin{document}

\maketitle

\begin{affiliations}
 \item Institute of Computing Technology(ICT), Chinese Academy of Sciences(CAS), Beijing, China.
 \item University of Chinese Academy of Sciences, Beijing, China.
 \item Xiyuan Hospital, China Academy of Chinese Medical Sciences(CACMS), Beijing, China.
 \item e-mail:lcui@ict.ac.cn.
\end{affiliations}

\begin{abstract}

Cervical spondylosis (CS) is a common chronic disease that affects up to two-thirds of the population and poses a serious burden on individuals and society. The early identification has significant value in improving cure rate and reducing costs. However, the pathology is complex, and the mild symptoms increase the difficulty of the diagnosis, especially in the early stage. Besides, the time-consuming and costliness of hospital medical service reduces the attention to the CS identification. Thus, a convenient, low-cost intelligent CS identification method is imperious demanded. In this paper, we present an intelligent method based on the deep learning to identify CS, using the surface electromyography (sEMG) signal. Faced with the complex, high dimensionality and weak usability of the sEMG signal, we proposed and developed a multi-channel EasiCSDeep algorithm based on the convolutional neural network, which consists of the feature extraction, spatial relationship representation and classification algorithm. To the best of our knowledge, this EasiCSDeep is the first effort to employ the deep learning and the sEMG data to identify CS. Compared with previous state-of-the-art algorithm, our algorithm achieves a significant improvement.

\end{abstract}

\maketitle
\section{Introduction}

\noindent
The cervical spondylosis(CS) is a common degenerative disease that harm human life and health, affects up to two-thirds of the population, and pose an serious burden on individuals and society~\cite{matz2009joint,kotil2008prospective,cai2016trend,wang2018convenient}.
The early identification is an effective means to improve the cure rate and reduce the cost.
However, the CS identification is difficult by analyzing the clinical symptom and the spinal lesions
for the following reasons:
(1) the CS is a complex disease associated with a series of clinical symptom\footnote{The clinical symptoms include neck and back pain, upper limb weakness, head
halo
nausea, vomiting, numbness of fingers, weakness of lower limbs, difficulty in walking, even
tachycardia, difficulty in swallowing, blurred vision, etc~\cite{Lufengyan2017, zhanshaoqun2016}.} and the spinal lesions\footnote{The spinal lesions include the vertebral bodies and intervertebral disks degeneration, disk ruptures, herniation, etc~\cite{wang2018convenient}.
Usually, as cervical degeneration worsens, clinical manifestations become more
obvious.}.
(2) there are many cases of cervical lesions in asymptomatic patients~\cite{liuzizhen2013}.
What's more, as the clinical symptoms and the spinal lesions of the early stages are relatively mild,
the difficulty of the diagnosis increase.
In addition, as the mild clinical symptoms of the early stages usually do not affect the live and work, it is easily overlooked by the suffers.
Meanwhile, the chronic degenerative disease CS will worsen over time if not intervened as early as possible.
Thus, the early identification is an essential, significant and an challenging task that need be paid more attention.

Currently, the clinical diagnosis mainly depend on the doctor's judgment from the clinical symptoms and  spinal lesions provided by expensive instruments in hospital~\cite{wang2018convenient}, which rely on the intervention of medical resources including the professionals and medical instruments.
However, the time-consuming and high cost of hospital medical service reduce people's attention to the early CS identification even if the suffers feel slightly uncomfortable in neck.
Meanwhile, the early identification is also hindered by the problem of difficulty and expensive to access medical service caused by the insufficient and uneven distribution of health care resources
considering the population~\cite{ChinaHealthandFamilyPlanningStatisticsYearbook,health2017,SixthNationalCensus,houxiaorong2014,anyanfang2011,
yingzhengxian2013}.
In detail, a large number of patients from all over the country flocked to the hospital with high-quality medical resources for medical service, exceeding the normal reception capacity of the hospital. The transportation, enrollment and queuing brings great inconvenience to the diagnosis and treatment of diseases, even delaying the diagnosis and treatment.
Besides, the expenditure is caused by the transportation and accommodation, further increasing the burden. With the development of the artificial intelligence(AI), it provides an opportunity for the people to seek a convenient, low-cost medical server. The artificial intelligence algorithm are used to realize the movable or even family of cervical spondylosis identification services, providing the convenient, low-cost service. As the current common means, the image-based intelligent CS identification attracts a lot of attention ~\cite{yu2015classifying,wang2018classification,wang2018convenient}.
Though it is the most intuitive, accurate and human resource-saving method at present,
the dependence on medical instruments resources makes this method inconvenience and costly.
A CS identification method is urgently needed, which enables people to conveniently and timely receive medical services at low cost to prompt the early identification and treatment of the CS.

\begin{figure*}
\centering
\epsfig{file = 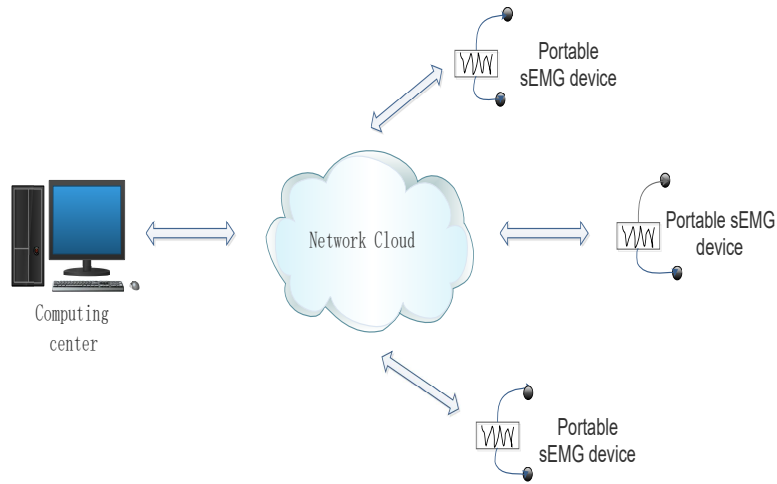, width=4.0in,height=2.5in}\\
\quad
\flushleft
\caption{The CS identification based on sEMG and machine learning.\label{fig: Figure 1}}
\quad
\end{figure*}

Fortunately, with the development of the biomedical electrode manufacturing technology~\cite{zhou2015development,lee2012comparison, zhang2010performance,merletti2009technology,kim2008influence,grajales2006wearable}, the acquisition of the sEMG signal become more portable and convenient.
And, as the CS-related physiological signals sEMG~\cite{johnston2008neck,madeleine2016effects,johnston2008alterations,falla2007muscle} is non-intrusive and affordable, it is a promising direction that use the sEMG
signal to provide the convenient, low-cost CS identification service as shown in the Figure~\ref{fig: Figure 1}.

The key of the harmless, convenient and low-cost CS identification is to classify subjects using sEMG signal.
However, the higher-accurate, higher-sensitivity and higher-specificity CS identification is an challenging task because of the following three folds:

\begin{itemize}
\item  The weak usability of the data:
       the instability of instruments\footnote{In addition to the Ag/AgCl electrode, the conductive gel is also used to collect data so that the large noise and errors are easily introduced~\cite{wang2018convenient}} and the long-term irregular movement habits of the subjects
       result in inaccurate measurement results, and the peeling of the electrode caused by the interference of the sweat lead the incompleteness of the single sample data.
       The inaccurate and incompleteness of the data directly affect the data quality, increase the difficulty of model's training, even affect the final classification performance and generalization performance.

\item High dimension and large dimension difference:
    the dimension of the sEMG sample are hundreds of thousands, the max, min and mean dimension of which respectively are $1453200$, $36000$, $328638$.
    The high dimension and the large dimension difference of the sEMG sample will increase computational and storage overhead and affect the generalization ability of the classification algorithm when they are directly handled by the machine learning models.

\item
    The complex spatial relationship: as the sEMG signals of the multiple muscles are related to the  recruitment of multiple muscles in the cervical spine system,
    there is complex spatial relation between the sEMG signals of the different muscles.
    The relationship above is difficulty to extract without the guidance of the expert knowledge and experience.

\end{itemize}

Currently, the performance of the CS identification using the sEMG signal is obtained by traditional machine learning model, the classification ability of which are bounded by the limited computational units of the shallow learning algorithm\footnote{The shallow learning algorithm refer to the traditional machine learning model, such as the support vector machine learning model (SVM), random forest (RF), the logistic regression (LR), boosting. The SVM, RF and boosting can be viewed structurally as the models with only one hidden layer node. The LR can be viewed structurally as the models without hidden layer node.} as well as the limited feature design capability.
Compared with traditional machine learning model, the deep learning are able to learn complex nonlinear function~\cite{zhou2017pdeep,lecun2015deep} by adding multiple hidden layers.
In addition, it has been shown that different layers of the deep neural network could learn different representations of objects automatically~\cite{zeiler2014visualizing,zhou2017pdeep}.
What's more, the convolutional neural network, an deep learning neural network, can extract the local relationship of the data and have lots of prior knowledge to compensate for the data we don't have~\cite{krizhevsky2014imagenet,jarrett2009best,krizhevsky2010convolutional}, improving the performance of classification and enhancing the generalization ability of the model.

In this paper, we developed a method EasiCSDeep based on the convolutional neural network to provide convenient, low-cost and intelligent CS identification with state-of-the-art performance, using sEMG signal.
The EasiCSDeep is a three-tier model: feature extraction, spatial relationship representation and classification algorithm.
For the feature extraction, we employed five kinds of feature extraction methods to extract features of the sEMG and characterize the high dimension sEMG signal as comprehensively as possible with the low dimensional features, effectively reducing data dimensions while preserving the raw sEMG signal properties.
For the spatial relationship representation, we reorganized the feature data into two-dimensional array data according to the anatomical location of the muscles that produces the sEMG signal as well as the type of activity involved, supporting the deep learning model to automatically extract relationship between the sEMG signals of the multiple muscles.
For the classification algorithm, we proposed a multi-channels model EasiDeep with six processing channels to process the different kinds of data simultaneously.
The EasiDeep retain the main features, reduce the number of parameters and automatically capture the dependency relationship without the domain knowledge, improving the performance of the model.
With the metric of accuracy, sensitivity, specificity and AUC, the EasiDeep achieves performance of 0.9708 in AUC, 97.22\% in accuracy, 100.00\% in sensitivity and 92.86\% in specificity.

Our contributions are the following five aspects:
\begin{itemize}
\item  To the best of our knowledge, this EasiCSDeep represents the first effort to employ the deep learning to analyze the sEMG data for intelligent CS identification.

\item
    We employed five types of feature extraction methods to extract five types of the features, each type feature focus on the properties of an aspect of the sEMG data, and characterize the sEMG signal as comprehensive as possible, reducing data dimensions while preserving raw sEMG signal properties. And each type of feature extraction method focus on the properties of the signals in a different observation view.

\item Inspired by the the First Law of Geography~\cite{tobler1970computer}, we reorganized the data combined the knowledge of the anatomy and kinematics, to improve the performance of the EasiCSDeep.

\item According to the feature types, the EasiDeep is developed to automatically extract the relationship without the expert knowledge and experience, and reduce the parameters of network while ensuring sufficient computing power of the network.

\item Compared with previous state-of-the-art models in the same classification task, our algorithm achieves a significant improvement in performance on same data set to further reduce personal misdiagnosis rate, overall misdiagnosis rate and missed diagnosis rate.
\end{itemize}

This work promotes the identification of physiological signal-based methods in a complex environment that is closer to the real application, and lays the foundation for the fine-grained classification of sEMG signals.

\begin{figure*}
\centering
\epsfig{file = 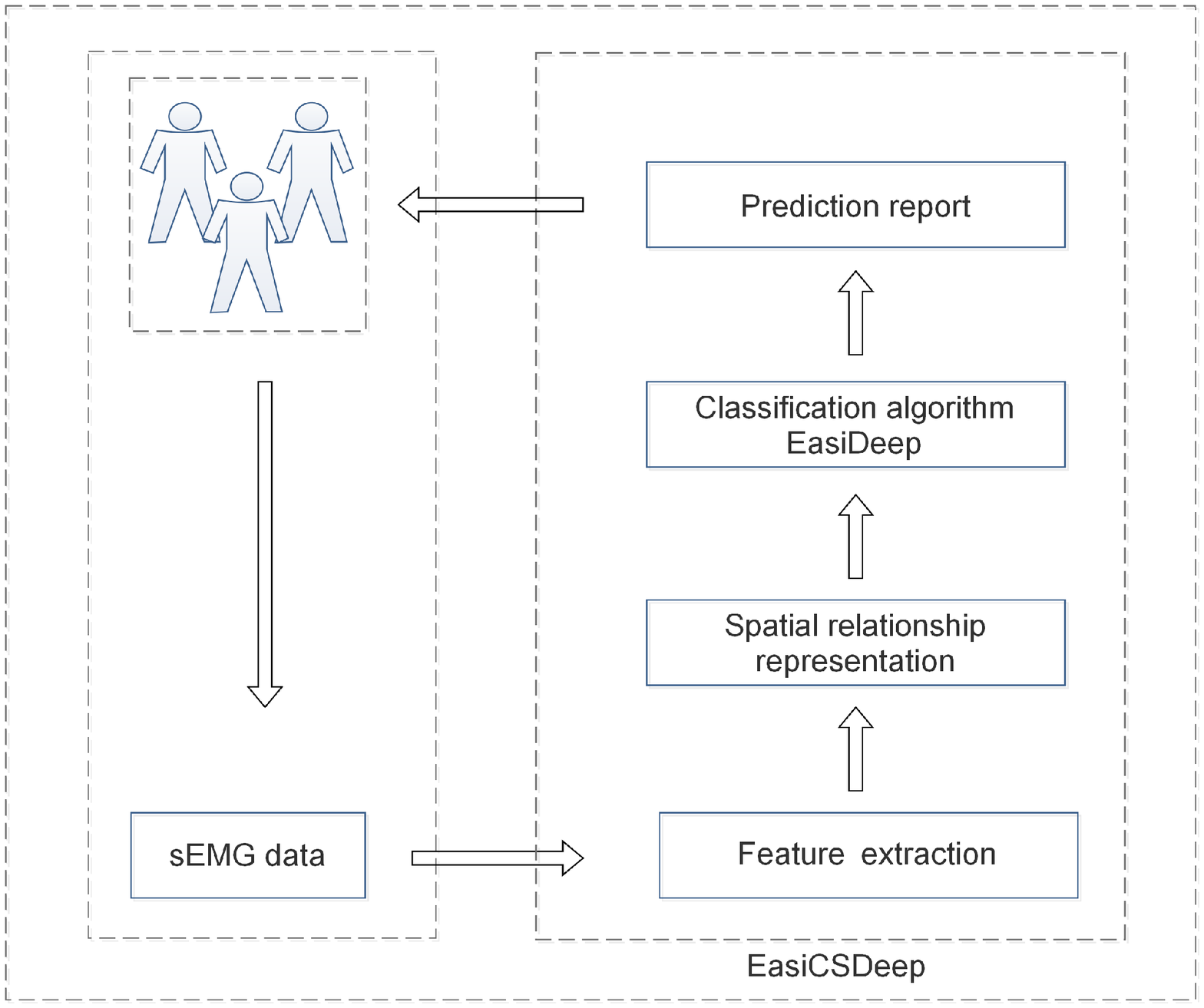, width=3.0in,height=2.8in}\\
\quad
\flushleft
\caption{The CS identification method based on the EasiCSDeep and sEMG signal.\label{fig: Figure 3}}
\quad
\end{figure*}

The paper is structured as follows.
Section 2 summarizes the CS identification methods and machine learning algorithms.
Section 3 describes the data set.
Section 4 format the CS identification tasks.
Section 5 introduces our proposed method EasiCSDeep.
Section 6 evaluates and discusses the results.
Section 7 concludes the paper and the future work.

\section{Related Work}

The CS, a chronic 'wear and tear' degenerative process of the cervical spine,
affects the physical and mental health of human beings and poses a serious economic burden
on society and individuals~\cite{matz2009joint,kotil2008prospective,cai2016trend}.
Currently, its diagnosis mainly depend on intervention of medical resources including the
professionals and medical instruments.
There are many study on the methods of CS intelligent identification: image-based method, clinical symptoms-based method and physiological-signal-based method.

As the common means, the image-based intelligent CS identification
attracted a lot of attention~\cite{yu2015classifying,wang2018classification,wang2018convenient}.
The research~\cite{yu2014classifying} extracted 10 effective ROIs to establish X-ray symptom-disease table of CS, and developed the fuzzy calculation model based on the table above to classify CS, yielding approximately 80.33\% accuracy. The research~\cite{yu2015classifying} put forward a method based on maximum likelihood theory to classify the types of cervical spondylosis using X-ray, achieving the accuracy of 80\% better than artificial X-ray reading method.
Combined with the diffusion tensor imaging metrics, the research~\cite{wang2018classification} develop a support vector machine(SVM) classifier to classify the groups with the cervical spondylotic from the control groups, achieving the accuracy of 95.73\%, sensitivity of 93.41\% and specificity of 98.64\%.
The image-based CS intelligent recognition is an accurate and human resource-saving method.
However, these minimal changes in the images in the early stage of the CS make visual diagnosis a difficult task and request more expert knowledge and experience. What's more,
the intervention of medical instrument resources make the method relative costly and inconvenient.

For the physiological-signal-based method, it is well known that the sEMG physiological signal
contains muscle lesion information,
and are widely utilized to study muscle function and status of the patients with neck disease~\cite{johnston2008neck,madeleine2016effects,johnston2008alterations,falla2007muscle}.
The research~\cite{johnston2008neck} utilize the sEMG to explore cervical musculoskeletal function in female office workers with neck pain, the result of which is consistent with that there are the altered muscle recruitment strategy in the female office workers with neck pain.
The research~\cite{falla2004patients} employ the sEMG data to compare neck muscle activation patterns during and after a repetitive upper limb task between patients with idiopathic neck pain, whiplash-associated disorders and the control group, demonstrating that patients with neck pain have greater activation of accessory neck muscles that represent an altered pattern of motor control to compensate for reduced activation of painful muscles.
The research~\cite{zhaozhongmin2011} use the linear model to study the stability and repeatability of the sEMG index MF and MPF of the cervical muscle in CS patients
as well as the healthy, and then employed the Student's t test to study the different of the index MF and MPF of the patients and the healthy,
which demonstrates the correlation between cervical muscle and CS.
However, there are few research on using the machine learning and sEMG signal to identify the CS.
The research~\cite{wang2018convenient} is the first work to identify CS using sEMG and AI.
It proposed the convenient and effective data collection method and the machine learning identification model EasiAI including the feature extraction, feature selection and classification algorithm, which construct a dataset involving 179 subjects from multiple type CS and achieve the 91.43\% in accuracy, 97.14\% in sensitivity, 81.43\% in specificity. It further demonstrates the feasibility of using  sEMG and AI for cervical spondylosis screening to a certain extent.
The performance of the EasiAI depend on feature engineering which requires expert knowledge and experience that lie in the deep and comprehensive understanding of the development of the CS.
However, the CS is an complex disease, every people is also an independent complex system, and the thorough and comprehensive exploration of the pathology of people's CS still need ongoing effort.
It becomes more difficult to improve the performance of the model by constructing more features without the guidance of the expert knowledge and experience above.

The implementation of CS intelligent identification system above depends on the powerful statistical analysis approaches machine learning.
The machine learning has been widely used in clinical data analysis to deal with the complicated data and has made great progress in medical~\cite{lezcano2017development, miotto2017deep, miotto2017deep, esteva2017dermatologist}.
It is roughly divided into traditional machine learning and deep learning.
When using traditional machine learning algorithms, we have to focus on
designing appropriate features manually in addition to the conventional feature extraction methods\cite{zeiler2014visualizing, zhou2017pdeep}.
The design of the features require the professional knowledge and experience, which is even an challenging
for domain experts.
The deep learning are representation-learning methods with multiple levels of representation, which allows to be fed with raw data and automatically discover the representations needed for detection or classification~\cite{miotto2017deep, lecun2015deep}.
And it has brought about breakthroughs in processing images, video, speech and audio~\cite{lecun2015deep}.

Although, the previous intelligent CS identification~\cite{wang2018convenient} obtained 91.43\% in accuracy, 97.14\% in sensitivity, 81.43\% in specificity.
There are still room in improvement.
It is essential and significant to reduce the overall misdiagnosis rate and missed diagnosis rate as much as possible for the population, and decrease the rate of misdiagnosis of individuals as much as possible.

In this paper, we seek to provide the convenient, low-cost and intelligent medical server of CS identification with state-of-the-art performance.
It is a promising study that utilize the convenience of sensor technology and intelligent algorithms of artificial intelligence to transfer high-quality medical services from hospitals to families.
It will promote the early detection and early treatment of the disease, reduce the cost and improve the cure rate. What's more, it will potentially alleviate the problem of difficulty and costliness to access
medical service that is caused by the imbalance in the distribution of population and medical resources in China to some extent.

\section{Dataset}

The data set are acquired from 179 subject, 109 of which are the CS patients and 70 of which are the healthy. The sEMG signal from 6 muscles, which includes the left sternocleidomastoid($M_0$), the left upper trapezius($M_1$), the left cervical erector spinae($M_2$), the right cervical erector spinae($M_3$), the right upper trapezius($M_4$) and the right sternocleidomastoid($M_5$), were synchronously recorded when the subject completed the movements in the order of the bow($A_0$), head backwards($A_1$), left flexion($A_2$), right flexion($A_3$), left rotation($A_4$), right rotation($A_5$), and hands up($A_6$), each movement of which is performed 3 times.

\begin{figure*}
\centering
\epsfig{file=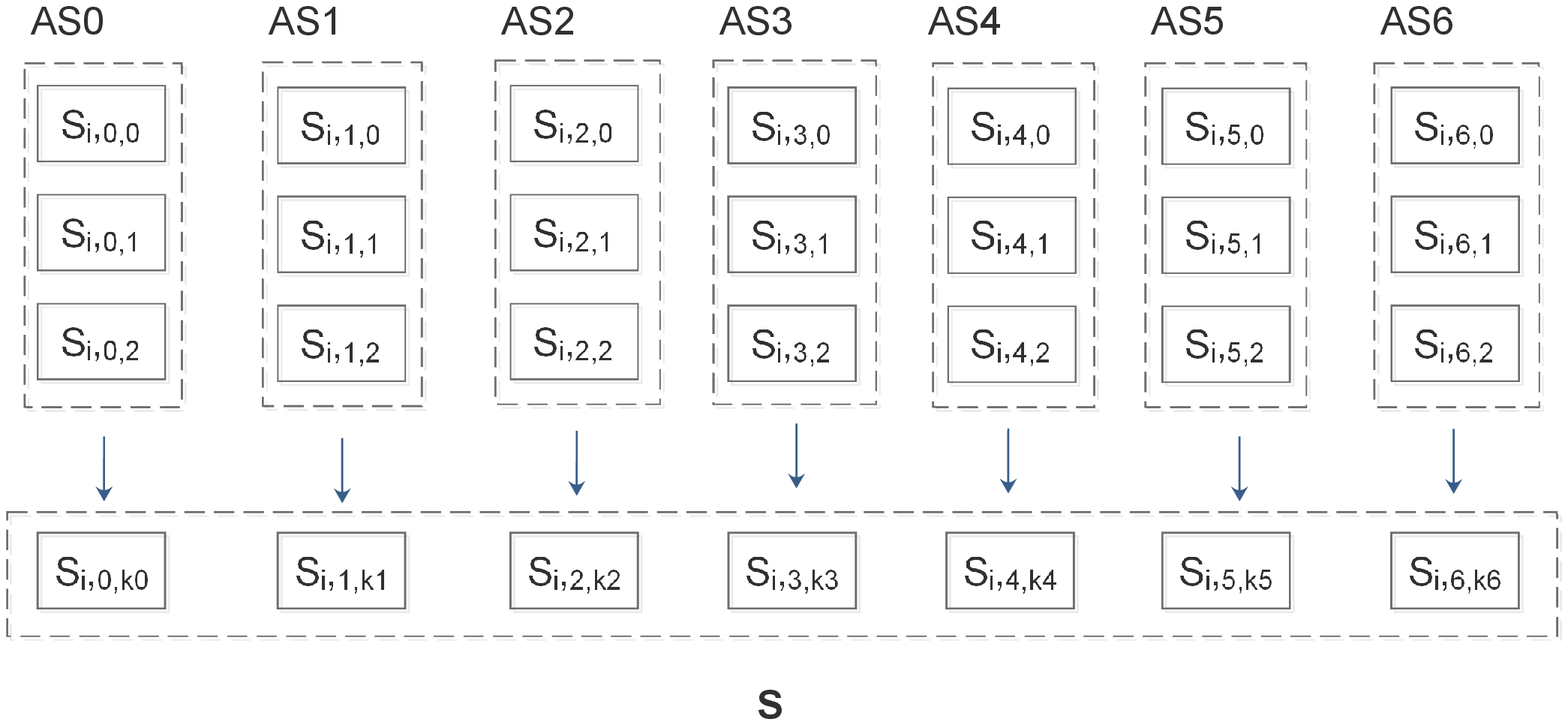, width=4.39in,height=2.00in}\\
\quad
\flushleft
\caption{The formation of a sample $S$ of a subject.
The $AS_i$ is the set of the $S_{i,j,k}$ from the movement $A_j$.
The $S_{i,j,k}$ denote the signal generated on the muscle $M_i$ $(0 \le i < 6)$ that is activated by the movement $A_j$ in $k-th$ time.  $0\le K0 < 3$, $0\le K1 < 3$, $0\le K2 < 3$, $0\le K3 < 3$, $0\le K4 < 3$, $0\le K5 < 3$. And the $K0$, $K1$, $K2$, $K3$, $K4$, $K5$ are respectively set to random integer between 0 and 3. \label{fig: Figure 2}}
\quad  
\end{figure*}

The movement $A_j$ that is performed in $k-th$ time is represented as $A_{j,k}$ which generate a $S_{i,j,k}$.
For a subject, the 3 $S_{i,j,k} (0 \le i < 6)$ is obtained from the movement $A_j$ and form a set that are denoted as $AS_j$.
As shown in the Figure~\ref{fig: Figure 2}, the $7$ $AS_j (0 \le j < 7)$, each $AS_j$ of which includes 3 $S_{i,j,k}(0 \le i < 6)$, is obtained from the 7 movement $A_j$.
And, respectively select 7 $S_{i,j,k}(0 \le i < 6)$ from 7 $AS_j$,
and splice the 7 $S_{i,j,k}(0 \le i < 6)$ in the order of $j=0$, $j=1$, $j=2$, $j=3$, $j=4$, $j=5$, $j=6$ to form a sample $S$. The 2187($1 \times 3^7$) samples are obtained from a subject, and
the 391473($179 \times 3^7$) samples are obtained from the 179 subjects.
The sample is represented as $S$ in the Formula~\ref{eq1}-~\ref{eq2}.

\begin{small}
\begin{align}\label{eq1}
S &\!=\!
\left[
  \begin{array}{cccccccc}
  S_{0,0} &S_{0,1} &S_{0,2} &S_{0,3} &S_{0,4} &S_{0,5} &S_{0,6}\\
  S_{1,0} &S_{1,1} &S_{1,2} &S_{1,3} &S_{1,4} &S_{1,5} &S_{1,6}\\
  S_{2,0} &S_{2,1} &S_{2,2} &S_{2,3} &S_{2,4} &S_{2,5} &S_{2,6}\\
  S_{3,0} &S_{3,1} &S_{3,2} &S_{3,3} &S_{3,4} &S_{3,5} &S_{3,6}\\
  S_{4,0} &S_{4,1} &S_{4,2} &S_{4,3} &S_{4,4} &S_{4,5} &S_{4,6}\\
  S_{5,0} &S_{5,1} &S_{5,2} &S_{5,3} &S_{5,4} &S_{5,5} &S_{5,6}\\
  \end{array}
\right],(0 \leq i < 6, 0 \leq j < 7)
\end{align}
\end{small}

The $i$ represent the muscle $M_i$, the $j$ represent the movement $A_j$. The $S_{i,j}$ denote the signal generated on the muscle $M_i$ in movement $A_j$.

\begin{small}
\begin{align}\label{eq2}
S_{i,j} &\!=\!
\left[
  \begin{array}{ccccccc}
  p_1 &p_2 &... &p_n\\
  \end{array}
\right],
\end{align}
\end{small}
Here, $p_n$ is a value converted from the electrical signal.

\section{The Formalization of the CS identification task}

In order to identify CS patients from the healthy, our task is described as follows.
The data set from the subjects are denoted as $D$.
Let $(X_{i}, y_{i}), (i\in{\{1,...N\}})$ denotes the $i$th sample in the data set $D$, where the $X_{i}$ is the sEMG data and the $y_{i}$ is the label with

\begin{small}
\[y_{i}=
\begin{cases}
0 &  \text{$X_{i}$ is sample of the healthy} \\
1 &  \text{$X_{i}$ is sample of the patient}
\end{cases}
\]
\end{small}

Our classification task is transformed to minimize the objective function $L(W)$ in the Formula \ref{eq3}.

\begin{small}
\begin{align}\label{eq3}
\centering
L{(W)} &\!=\! \sum_{i=1}^n{loss1(X_{i}, y_{i}; W)}+ \sum_{i=1}^n{loss2(X_{i}, y_{i}; W)} +F(W)
\end{align}
\end{small}

The $n$ denotes the number of the samples and the $W$  is the parameters of the model.
We employed the softmax function as the last output layer and the softmax cross-entropy function $loss1(X_{i}, y_{i}; W)$ as the loss function so that the health status and illness status can be independently optimized.
Let $p_j(X_{i}, y_{i}; W)$($ j \in \left \{0, 1 \right \}$) denotes the $j$th input of the softmax layer, then the $j$th activation of the softmax layer is $y_j(X_{i}, y_{i}; W)$ in the Formula ~\ref{eq4}.

\begin{equation}\label{eq4}
\centering
\small
y_j(X_{i}, y_{i}; W) \!=\! \frac{e^{p_j}}{\sum_{i=1}^j{e^{p_i}}}
\end{equation}

The softmax cross-entropy function $loss1(X_{i}, y_{i}; W)$ is shown in Formula~\ref{eq5},

\begin{equation}\label{eq5}
\centering
\small
loss1(X_{i}, y_{i}; W) \!=\! -{\widehat{y}_{i}} \times \sum_{i=1}^n{\times{\ln{y_{i}}}}
\end{equation}

where the $\widehat{y}_{i}$ is the true value and $y_{i}$ is the prediction value.
The $\widehat{y}_{i}$, the value of which is $0$ or 1 at the same time, make sure that only one label is optimized independently.
In order to make the predicted value $y_{i}$ as close as possible to the true value $\widehat{y}_{i}$, we simultaneously minimize the loss function $loss2(X_{i}, y_{i}; W)$ in Formula~\ref{eq6}.
In addition, we employed the classic $L_2$ regularized loss function $F(W)$ to counter overfitting in Formula \ref{eq7},

\begin{equation}\label{eq6}
\centering
\small
loss2(X_{i}, y_{i}; W) \!=\! \sum_{i=1}^n{(y_{i}-{\widehat{y}_{i}})^2}
\end{equation}

\begin{equation}\label{eq7}
\centering
\small
F(W) \!=\! {\alpha} \times{\left \| W \right \|_{2}^2}
\end{equation}
where the $\alpha$ is the regularization parameter.
We divide the train data set, validation data set and test data set by the ratio of 3:1:1 and make sure that the sample from a subject only appears in one data set.

\section{The Methodology}

\begin{figure*}
\centering
\epsfig{file=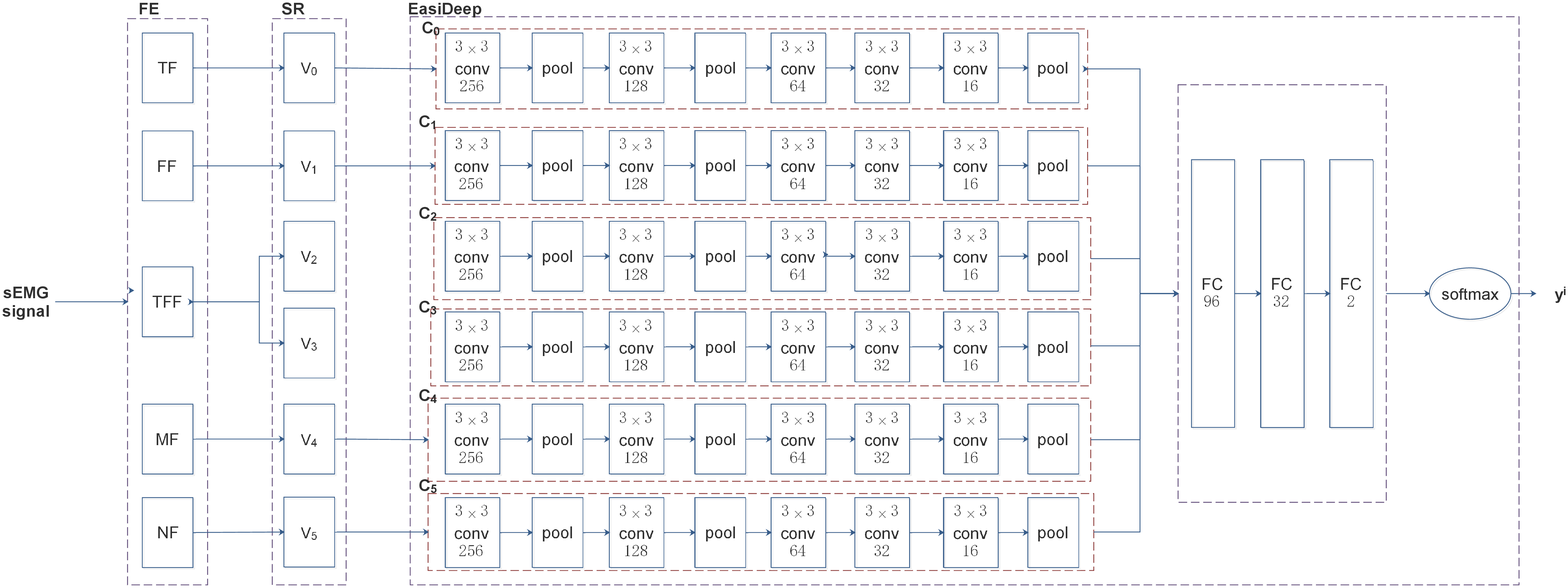, width=6.0in, height=2.6in}\\
\quad
\flushleft
\caption{The EasiCSDeep.\label{fig: Figure 4}}
\quad
\end{figure*}

In this section, we elaborate our proposed EasiCSDeep framework.
As shown in Figure~\ref{fig: Figure 4}, the EasiCSDeep consists of the feature extraction $FE$, spatial relationship representation $SR$ and classification algorithm EasiDeep.

\subsection{The Feature extraction}

Faced with the high dimensionality and weak usability of the sEMG data,
five types feature extraction methods, which includes
time-domain methods ($TM$), frequent-domain methods ($FM$),
the time-frequency method ($TFM$), the model-based methods ($MM$) and nonlinear entropy feature extraction method ($NM$), are used to extract the features and characterize the sEMG signal. Each type of the feature extraction method focus on the properties of the signals in the different observation view.

\subsection{Time-domain method}

The sEMG data is treated as the physical signal values on the time axis in time domian,
the variation property of which is usually represented by the statistical feature such as the the average, variance, maximum and minimum etc.
The time-domain methods are used to extract the feature above,
which consider the sEMG signal as a random signal, the mean value of which is zero and the variance of which varies with the intensity of the signal.
In this paper, we extracted 11 time-domain features from the sEMG signal $S_{i,j}$, forming features set $TF_{i,j}$.
The $TF_{i,j}$ includes $mean$, $var$(variance), $std$(standard deviation), $mode$(the most common value), $max$(maximum value), $min$(minimum value), $over\_zero$, $range$, $aemg$(averaged electromyogram), $iemg$(integrated electromyogram), $rms$(root mean square).
The 11 types of features are extracted by the 11 calculation methods.

\begin{small}
\begin{align}
mean &\!=\! 1/n \sum_{i=1}^n{p_i}\\
range &\!=\! max-min\\
iemg &\!=\! \sum_{i=1}^n{|p_i-mean|}\\
aemg &\!=\! 1/n \sum_{i=1}^n{|p_i-mean|}\\
rms  &\!=\! 1/n \sum_{i=1}^n{p_i^2}
\end{align}
\end{small}

\subsection{Frequency domain method}
The frequency-domain analysis method focus on variation characteristics of the myoelectric signals in the frequency domain,
the characteristics of which can be denoted as the frequency-domain features.
In this paper, we utilized the fourier transform to convert the sEMG signal $S_{i,j}$ into $SF_{i,j}$, and extracted 14 frequency-domain feature values:
$dc$(direct current), $mean$, $var$, $std$,
$skew$(skewness), $kurt$(kurtosis), $entropy$,
$s\_mean$(shape
mean), $s\_std$(shape standard deviation), $s\_var$(shape variance),
$s\_skew$(shape skewness), $s\_kurt$(shape kurtosis), $mf$(median frequency), $mpf$(mean power frequency).
The frequency domain feature set $FF_{i, j}$ are obtained.
\begin{small}
\begin{align}
SF_{i,j} &\!=\!
\left[
  \begin{array}{ccc}
  x_0,  &x_1, ..., &x_{k-1}\\
  \end{array}
\right],\\
a &\!=\! int(k/2) +1\\
dc &\!=\! x_0 \\
skew &\!=\! \frac{1}{a}\sum_{i=1}^{a}{((x_i - mean)/std)^3}\\
kurt &\!=\! \frac{1}{a}\sum_{i=1}^{a}{((x_i - mean)/std)^4 - 3}\\
s\_mean &\!=\! \frac{\sum_{i=1}^{a}{i x_i}}{\sum_{i=1}^{a}{x_i}}
\end{align}
\end{small}

\subsection{Time-frequency domain method}
Unlike the time domain and the frequency domain, the time-frequency domain method focus on both the time variation characteristic and frequency distribution simultaneously.
Different from the features in time domain or frequency domain,
the sEMG are able to be represented by the wavelet coefficients
generated in the process of the wavelet transform~\cite{acharya2017automated}.
In this paper, we firstly use the famous method discrete wavelet transform(DWT) to decompose sEMG signal at 5 levels.
Then, we extracted the approximation coefficients $C_h$ $\footnote{only the coefficients which is large than 0 will be extracted}$ at level 5.
Meanwhile, we respectively extracted the detail coefficients$ C_{l,5}$, $C_{l,4}$,
$C_{l,3}$, $C_{l,2}$ \footnote{only the coefficients which is large than 0 will be extracted} at the level 5, 4, 3, 2.
Finally, we extracted the feature $c_{max}$,
singular value, and $c_{energy}$ from the $C_h$, $C_{l,5}$, $C_{l,4}$, $C_{l,3}$,
$C_{l,2}$ respectively.
Each sEMG signal $S_{i,j}$ can be represent by 15 features
above. The $TFF\_{WT}\_{i, j}$ are obtained from $S_{i,j}$.

\begin{small}
\begin{align}
c_{max} &\!=\! Max  \left [ \log_{10}(x), x \in C \right]\\
C_{energy} &\!=\! \log_{10} \frac {\sqrt {\sum_{x \in C}{x^2}}}{|C|}
\end{align}
\end{small}

In addition, the signal is also decomposed into the
coefficients by the wavelet packet decomposition(WPD) when feed into the low (L) and high (H) filter at the same time\cite{alakus2017detection} in the Equation \ref{eq23}.

\begin{small}
\begin{align}\label{eq23}
\psi_{j,k}^i(t) &\!=\! 2^{-j/2}\psi^i(2^{-j}t-k), \quad (i=1,2, ..., j^n)\\
\psi^{2i}(t) &\!=\! \frac{1}{\sqrt{2}}\sum_{k=-\infty}^{\infty}h(k)\psi^i(\frac{t}{2}-k)\\
\psi^{2i+1} &\!=\! \frac{1}{\sqrt{2}}\sum_{k=-\infty}^{\infty}g(k)\psi^i(\frac{t}{2}-k)
\end{align}
\end{small}

Here, $i$ is the modulation parameter, $j$ is the dilation parameter, $k$ is the translation parameter, $n$ is the level of the decomposition, $h(k)$ and $g(k)$ are the quadrature mirror filters\cite{shinde2005wavelet}.

\begin{equation}\label{eq26}
\centering
\small
c_{j,k}^i=\int_{-\infty}^{\infty}f(t)\psi_{j,k}^i(t)dt
\end{equation}

As shown in the Formula \ref{eq26}, the $c_{j,k}^i$ is the coefficients of the WPD, and $f(t)$ is the signal to
be decomposed~\cite{shinde2005wavelet}.
In this paper, we set the level of the decomposition to $n = 3$ and each sEMG signal is represented by
8 coefficients.
After the WPD coefficients extraction, we obtained the feature set $TFF\_{WP}\_{i, j}$ from $S_{i,j}$.

\subsection{The model-based method}

For the model-based methods, we focus on the character of the data joint distribution.
Usually, the $x+1$-th value of the sEMG have a relationship with the previous signals, and the relationship above is represented by the coefficient of the model.
The Autoregressive (AR) model, a popular feature extraction method, is widely applied for the
physiological signals analysis
to characterize the joint distribution of the sEMG by representing the sEMG signal as the coefficients of the
model~\cite{koccer2017classifying} as shown in the Equation~\ref{eq12}.

\begin{small}
\begin{align}\label{eq12}
y(t) &\!=\! \sum_{i=1}^p{\phi_i y(t-i) + \varepsilon_i}
\end{align}
\end{small}

Here, $y(t)$ is the signal to be modeled, and $\phi_i$ are the model coefficients,
$\varepsilon_i$ is white noise.
The $p$, which determines the
number of the coefficients and the complex of the model,
is the order of the AR model.
We set the $p = 10$ and
$p = 4$, and train two AR model by the sEMG signal respectively.
Thus, each sEMG signal $S_{i,j}$ is represented as $MF_{i, j}$ including 14 RA
model coefficients above.

\subsection{Nonlinear feature extraction method}

As the sEMG signals are generated by the exchange of ions across the muscle membranes, the states of the exchange of ions are detected by the electrodes~\cite{altin2016comparison}, and the change of the states reflect the degree of mess of the muscle's activity.
We use entropy to describe the degree of the mess of the muscle's activity.
In detail, each value of the sEMG signal represent a
state, and the entropy is denoted as Equation~\ref{eq13}.

\begin{small}
\begin{align}\label{eq13}
entropy &\!=\! \sum_{i=1}^n{p_i \log(p_i)}
\end{align}
\end{small}

Here, the $n$ is the number of the states in the system, and $p_i$ is the
probability of the state $i$ in the system.
The set $NF_{i, j}$ are generated after we extracted the $entropy$ index from the sEMG signal $S_{i,j}$.

\subsection{Spatial relationship representation}

In this paper, we obtained the six kinds of features set: $TF$
\footnote{
The $TF$ includes $6 \times 7 \times 11$ $TF_{i,j}$.},
$FF$\footnote{The $FVF$ includes $6 \times 7 \times 14$ $FF_{i,j}$.},
$TFF_{WT}$\footnote{The $TFF_{WT}$ includes $6 \times 7 \times 15$ $TFF\_{WT}\_{i,j}$.},
$TFF_{WP}$\footnote{The $TFF_{WP}$ includes $6 \times 7 \times 8$ $TFF\_{WP}\_{i,j}$.},
$MF$\footnote{The $MF$ includes $6 \times 7 \times 14$ $MF_{i,j}$.},
$NF$\footnote{The $NF$ includes $6 \times 7 \times 1$ $NF_{i,j}$} from a sample including $6 \times 7$ sEMG signals $S_{i,j}$.
For the convenience of introduction, we respectively represented $TF$, $FF$, $TFF_{WT}$, $TFF_{WP}$,
$RF$, and $SF$ as $F_0$, $F_1$, $F_2$, $F_3$, $F_4$, $F_5$.

Inspired by that the First Law of Geography~\cite{tobler1970computer}, using the regions with strong correlation to predict the CS can improve the performance of the machine learning algorithms~\cite{yao2018deep}.
What's more, considering the relationship between the muscles activated by the same movement,
we reshape the view features data $F_k$ into the $V_k$, and the $V_k$ meets the following requirements:
(1)the data on the muscles involved in the same movement are put together.
(2)the data on the muscles whose physiological position are closer are put together closer.

As shown in Formula \ref{eq8}, we reshape the features $F_k$ into $V_k$.
The $i$th row of the $V_k$ are forming by stitching the $F_{i,j}$ in the order of the $j$ from $0$ to $6$.
The $j$th column are forming by splicing the $F_{0,j}$, $F_{5,j}$, $F_{2,j}$, $F_{3,j}$, $F_{1,j}$ and $F_{4,j}$ in turn.

\begin{small}
\begin{align}\label{eq8}
V_k &\!=\!
\left[
  \begin{array}{ccccccc}
  F_{0,0}, &F_{0,1}, &F_{0,2}, &F_{0,3}, &F_{0,4}, &F_{0,5}, &F_{0,6}\\
  F_{5,0}, &F_{5,1}, &F_{5,2}, &F_{5,3}, &F_{5,4}, &F_{5,5}, &F_{5,6}\\
  F_{2,0}, &F_{2,1}, &F_{2,2}, &F_{2,3}, &F_{2,4}, &F_{2,5}, &F_{2,6}\\
  F_{3,0}, &F_{3,1}, &F_{3,2}, &F_{3,3}, &F_{3,4}, &F_{3,5}, &F_{3,6}\\
  F_{1,0}, &F_{1,1}, &F_{1,2}, &F_{1,3}, &F_{1,4}, &F_{1,5}, &F_{1,6}\\
  F_{4,0}, &F_{4,1}, &F_{4,2}, &F_{4,3}, &F_{4,4}, &F_{4,5}, &F_{4,6}\\
  \end{array}
\right],
\end{align}
\end{small}

\subsection{The EasiDeep}

For each kind of data $V_i$, we employed channels $C_i$ to process it.
The channel consists of the 5 convolution layers and 3 pooling layers.
The transformation at each convolution layer $k$ is defined as follows:
\begin{equation}\label{eq14}
\centering
\small
Y^{k} = f(Y^{k-1} \times W^{k} + b^{k})
\end{equation}
The $Y^{k}$ is the output of the $k$-th layer, the $*$ denotes the convolutional operation, the $f$ is the activation function. In this paper, we use the Relu function as the activation function.
The $W^{k}$ and $b^{k}$ respectively represent parameters and biases of the convolution layer $k$.
The transformation at each pooling layers $k$ is defined as follows:
%pool层
\begin{equation}\label{eq15}
\centering
\small
Y^{i} = g(Y^{i-1})
\end{equation}

The $g$ is the pooling function. In this paper, we use the max function as the pooling function.
After the channels, the outputs $Y^{k}$ are flatten and stitched into a feature vector $FY^{k}$ ($FY^{k} \in R^{S_i \times S_j \times 1}$), then feed into the two fully connected
layers to reduce the dimension, each fully connected layer $i$ of which is defined as:
%全连接层
\begin{equation}\label{eq16}
\centering
\small
FY^{(i)} = f(FY^{i-1}\times W^{i} + b^{i})
\end{equation}
The $W^{i}$ and $b^{i}$ respectively are the parameter and biases of the fully connected layer $i$.
Finally, the output of the last fully connected layer is feed into the softmax layer.

%softmax
\begin{equation}\label{eq17}
\centering
\small
Y^{(i)} = softmax(FY^{i-1}\times W^{i} + b^{i})
\end{equation}

The $Y^{(i)}$ is a vector $(y_0^i, y_1^i)$. The $y_0^i$ and $y_1^i$ are greater than 0 and is less than 1, and
the sum of $y_0^i$ and $y_1^i$ is equal to 1.
The $y_0^i$ denotes the probability that the sample is the healthy.
The $y_1^i$ denotes the probability that the sample is the CS suffer.

\subsection{Algorithm and Optimization}

As the Algorithm ~\ref{alg:alg1} shown, we firstly extract the features
from the raw sEMG in different views respectively. Then,
the EasiCSDeep is trained by the back-propagation and the AdamOptimizer.

\begin{algorithm}
    \renewcommand{\algorithmicrequire}{\textbf{Input:}}
    \renewcommand{\algorithmicensure}{\textbf{Output:}}
    \caption{EasiDeep Training Algorithm}
    \label{alg:alg1}
    \begin{algorithmic}[1]
    \REQUIRE The sEMG of samples: \{$X_0, X_1, ..., X_{n-1}$\};\\
                      \quad \ \ The label of the samples:\{$y_0, y_1, ..., y_{n-1}$\}.
    \ENSURE Learned EasiCSDeep model
    \STATE //construct training instances
    \STATE $D \leftarrow \phi$
    \FOR {all sEMG of samples}
    \STATE the $TF$ features in the $TM$ method\;
    \STATE extracting the features $V_1$ in the $FV$ view.
    \STATE extract the $FF$ features in the $FM$ method\;
    \STATE extract the $TFF_{WT}$ features in the $TFM$ method\;
    \STATE extract the $TFF_{WP}$ features in the $TFM$ method\;
    \STATE extract the $NF$ features in the $NM$ method\;
    \STATE represent \{$TF$, $FF$, $TFF_{WT}$, $TFF_{WP}$, $MF$, $NF$\} into \{$V_0$, $V_1$, $V_2$, $V_3$, $V_4$, $V_5$\}\;
    \STATE //y is the label of the sample
    \STATE put an training instance (\{$V_0$, $V_1$, $V_2$, $V_3$, $V_4$, $V_5$\}, y) into $D$\;
    \ENDFOR
    \STATE // train the model
    \STATE initialize all learnable parameters $\theta$ EasiDeep
    \REPEAT
    \STATE select a batch of instances $D_b$ from $D$
    \STATE find $\theta$ by minimizing the Formula \ref{eq3} with $D_b$
    \UNTIL{stopping criteria is met}
    \end{algorithmic}
\end{algorithm}

\section{Experiments Results, Evaluation and Discussion}

With the metric of accuracy, sensitivity, specificity and AUC(the area under the sensitivity
and specificity curve)~\cite{esteva2017dermatologist,baldi2014searching}, we validated the
effectiveness of the EasiCSDeep in three ways: (a)compare the EasiCSDeep with other machine learning models.(b) compare the EasiCSDeep with the various configurations of the channel combination.(c)compare the EasiCSDeep with the various configurations of filter size.

\subsection{The metrics}

The good intelligent CS identification model primarily are able to decrease the rate of misdiagnosis of individuals as much as possible, and reduce the overall misdiagnosis rate and missed diagnosis rate as much as possible for the population.
The higher accuracy is, the lower the rate of misdiagnosis of individuals.
The higher sensitivity and specificity, the lower the overall missed diagnosis rate and misdiagnosis of the population are.
So we use the accuracy, sensitivity and specificity as the metrics in our paper, and they are defined as follows.
\begin{small}
\begin{align}
accuracy &\!=\! \frac {tp + tn}{p + n}\\
sensitivity &\!=\! \frac {tp}{p}\\
specificity &\!=\! \frac{tn}{n}
\end{align}
\end{small}
where the 'tp' is the number of correctly predicted patient with CS, ’p’ is the
number of patient with CS shown, 'tn' is the number of correctly predicted healthy
people free of CS, and the 'n' is the number of healthy free of CS shown.
The higher the sensitivity is, the lower the missed diagnosis rate is. The higher the specificity is, the lower the misdiagnosis rate is. Furthermore, the high sensitivity boosts the early detection of the more patient population. The higher specificity decreases the misdiagnosis rate.

\subsection{Models for Comparison}

We compared the EasiCSDeep with the traditional machine learning models: logistic regression(LR), support vector machine(SVM), random forest(RF) and EasiAI on the same data set in the same classification task.
In addition, we also considered the comparison with a typical deep learning
model $DL_{cnn}$ based on CNN which was developed in our paper.
The details on the models above is as follows:
\begin{itemize}
\item LR: The logistic regression is a classic linear regression method. The output of the LR is between 0 and 1.
          The closer the output value is to 1, the more likely the sample is positive.
\item SVM: The SVM is a classic two-class model. It can find the optimal hyperplane of the n-dimensional space to separate the positive and negative samples.
\item RF: The RF is a classic bagging-based ensemble learning algorithm, which consists of multiple decision trees.
          The prediction result is voted by all the decision trees.
\item EasiAI~\cite{wang2018convenient}: The EasiAI is an ensemble learning algorithm based on the GBDT(gradient boosted regression tree), which includes the feature extraction, feature selection and classification algorithm.
\item $DL_{cnn}$: The architecture of $DL_{cnn}$ is the same as $C_i$.
        The $DL_{cnn}$ consists of the 5 convolutional layers, 3 pooling layers, 3 fully connected layers and 1 softmax function.
\end{itemize}

\subsection{Preprocessing and Parameters}

We used the mean of the features to fill in the missing data, and
employed the StandarScaler provided by the sklearn.processing ($0.19.1$) to process the data set.
In addition, the label is transformed into the two-dimensional array, using the one-hot encoding.

All the experiments in the paper were run on the cluster with four NVIDIA Corporation GP102 [TITAN Xp] (rev a1).
The python ($3.6.5$) and tensorflow ($1.1.0$) were used to implement the deep learning algorithmes in the paper.
The number of the convolution layer is $5$. The pooling layer number is $3$, and the fully connected layer number is 3.
The activation function for each layer is Relu, and
the batch normalization is used to process the input of the Relu.
The filter size of each convolution layer is $5 \times 5$. The kernel size of pooling layer is $2 \times 2$.
The kernel number of the convolution layers respectively are 256, 128, 64, 32, 16.
The number of neural units in the fully connected layer respectively is 96, 32, 2.
The batch size is set to 148, and the learning rate is 0.00006.
The max epoch is 10000.
The 60\% of the samples is used for training the model, 20\% is used as the validation data set for selecting the model
with the best performance.
The early-stop was also applied when training the model, the round of which is less than 1500 in all the experiments.

\subsection{Performance Comparison}

\subsubsection{Comparison with the other machine learning models}

As detailed in Table~\ref{tab:tab1},
the EasiCSDeep achieved the highest AUC 0.9708, the highest accuracy 97.22\%, the highest
sensitivity 100\% and the highest specificity 92.86\%.
We compared the EasiCSDeep with the the traditional machine learning model of LR, SVM, RF and EasiAI.
Compared with the LR with the lowest sensitivity, the EasiCSDeep obtained 10.65, 11.08, 10.00 and 0.0508 percentage point improvement in accuracy, sensitivity, specificity and AUC.
Compared with the SVM with the lowest accuracy, the EasiCSDeep got 11.22, 10.22, 12.86 and 0.0508 percentage point improvement in accuracy, sensitivity, specificity and AUC.
Compared with the RF with the lowest specificity, the accuracy, sensitivity, specificity and AUC has been improved by 9.57, 6.58, 14.29 and 0.0308 percentage point.
Compared with the previous state-of-the-art identification model EasiAI, the EasiCSDeep obtained 6.2, 2.86, 11.43 and 0.0208 percentage point improvement in classification accuracy,
sensitivity, specificity and AUC.

In addition, we compared the EasiCSDeep with the deep learning model $DL_{cnn}$.
The EasiCSDeep obtained 5.79\%, 4.75\%, 8.7\%, 0.0014 improvement in the accuracy, sensitivity, specificity and AUC.
Overall, the EasiCSDeep provides the state-of-the-art classification performance, proving that the deep learning algorithm outperform the traditional machine learning.
Multi-channel models (EasiCSDeep) have better classification performance than single-channel model($DL_{cnn}$).

\begin{table}
%\small
\caption{The performance comparison between the EasiCSDeep and previous research.\label{tab:tab1}}
\centering
\begin{tabular}{ccccc}
\hline
Model      &AUC       &Accuracy&Sensitivity&Specificity\\
\hline
LR         &0.9200    &86.57\% & 88.92\%   &82.86\%\\
SVM        &0.9200    &86.00\% & 89.78\%   &80.00\%\\
RF         &0.9400    &87.65\% & 93.42\%   &78.57\%\\
EasiAI     &0.9500    &91.02\% & 97.14\%   &81.43\%\\
$DL_{cnn}$ &0.9694    &91.43\% & 95.25\%   &84.16\%\\
EasiCSDeep &0.9708    &97.22\% & 100.00\%  &92.86\%\\
\hline
\end{tabular}
\end{table}

\subsubsection{Comparison of the EasiCSDeep
with the various channel combination}

We evaluate the EasiCSDeep with all the configurations of the channel combination.
The channel $C_0$, $C_1$, $C_2$, which are fed with the feature data extracted in the commonly methods,
are considered as the basis channels.
Combine the basic channels and other channels $C_3$, $C_4$ and $C_5$, and compare the performance of the model with different channel combination.
The 8 models with various channel combination are shown in the Table~\ref{tab:tab2}.
In the three-channels model with the combination of channel $C_0$, $C_1$, $C_2$, the accuracy is 78.25\%.
In the three-channels model with the combination of channel $C_0$, $C_1$, $C_2$, $C_3$, the accuracy is 83.33\%.
In the five-channels model with the combination of channel $C_0$, $C_1$, $C_2$, $C_4$, $C_5$ the accuracy is 86.11\%.
In the six-channels model with the combination of channel $C_0$, $C_1$, $C_2$, $C_3$, $C_4$, $C_5$ the accuracy is 97.22\%.
It is concluded that the addition of the channel $C_3$ improve the accuracy to a certain extent.
In all the four-channels models, the highest accuracy is 86.11\%, the lowest accuracy is 80.00\% and
the mean accuracy is 83.15\% which is large than 78.25\% of three-channel model.
In all the five-channels models, the highest accuracy is 94.44\%, the lowest accuracy is 86.11\% and
the mean accuracy is 89.81\% which is large than the mean accuracy 83.15\% of the four-channels models.
The EasiCSDeep with six channels provides the highest accuracy 97.22\%.
Overall, the highest and lowest accuracy rate rise as the number of the channels increases.
The results validate the effectiveness of the six-channels EasiCSDeep.
As the different channel is almost used to process a type of the feature data, which focus on the property of one aspect of sEMG signal,
it is also indicated that the sEMG signal of CS is an complex data which have the variation characters.

\begin{table}
\caption{The comparison between different EasiCSDeep models with different channel combination.\label{tab:tab2}}
\centering
\begin{tabular}{cccccccc}
\hline
Index &$C_0$   &$C_1$  &$C_2$  &$C_3$ &$C_4$ &$C_5$ &Accuracy\\
\hline

0&  \checkmark &\checkmark &\checkmark &&&                         &78.25\%\\
1&  \checkmark &\checkmark &\checkmark &\checkmark&&               &83.33\%\\
2&  \checkmark &\checkmark &\checkmark &&\checkmark&               &86.11\%\\
3&  \checkmark &\checkmark &\checkmark &&&\checkmark               &80.00\%\\

4&  \checkmark &\checkmark &\checkmark &\checkmark &\checkmark &   &94.44\%\\
5&  \checkmark &\checkmark &\checkmark &\checkmark & &\checkmark   &88.89\%\\
6&  \checkmark &\checkmark &\checkmark & &\checkmark &\checkmark   &86.11\%\\
7&  \checkmark &\checkmark &\checkmark &\checkmark &\checkmark &\checkmark  &97.22\%\\
\hline
\end{tabular}%}
\end{table}

\subsubsection{Comparison of the EasiCSDeep
with the various filter size}
We assess our model EasiCSDeep with the various configurations of filter size in the Table ~\ref{tab:tab3}.
As detailed in Table~\ref{tab:tab3}, the EasiCSDeep obtain the highest accuracy 97.22\% when the filter size is $(5, 5)$.
When the filter size is $(2, 2)$, the accuracy is lowest 83.67\%.
In general, the accuracy rises as the filter size increases.
It concluded that the spatial relationship representation has an positive effect on the CS classification task, and
it demonstrated that using the regions with strong correlation to predict
the CS can improve the performance of the machine learning algorithms~\cite{yao2018deep}.

In turn, two conclusions are summarized: (1)there are strong relationship between the sEMG signal of the six muscles activated by the same movement.(2)there are strong relationship between the sEMG signal of the muslces whose physiological position are closer.
It can be further explained by that:(1) the sEMG of the six muscles are affected by the muscles recruitment pattern. (2) there are differences in muscle recruitment patterns between the CS and the healthy.
%(3)the occurrence of CS is accompanied by changes in muscle recruitment patterns, the change of which appear in patients with neck pain~\cite{johnston2008alterations, madeleine2016effects, johnston2008neck, falla2004patients, falla2004unravelling, falla2007muscle}.
%(2) the muscles of the neck are present in a system and play different roles and work together to complete the basic functions of the neck.
%it is concluded that there are differences in muscle recruitment patterns and the occurrence of CS is accompanied by changes in muscle recruitment patterns, the change of which appear in patients with neck pain~\cite{johnston2008alterations, madeleine2016effects, johnston2008neck, falla2004patients, falla2004unravelling, falla2007muscle}.

\begin{table}
\small
\caption{The comparison between different EasiCSDeep models with different configuration of filter size.\label{tab:tab3}}
\centering
\begin{tabular}{c|ccc} %{p{1cm}p{1cm}p{1cm}p{1cm}p{1cm}p{1cm}} %{cccccccc}
\hline
Channels &Filter size &Pool size &Accuracy\\
\hline
6 &(6,6)       &(2,2)     &86.11\%\\
6 &(5,5)       &(2,2)     &97.22\%\\
6 &(4,4)       &(2,2)     &88.89\%\\
6 &(3,3)       &(2,2)     &82.13\%\\
6 &(2,2)       &(2,2)     &80.56\%\\
\hline
\end{tabular} %}
\end{table}

\section{Conclusions and future work}

In this paper, we have proposed and developed an novel multi-channel
model EasiCSDeep based on deep learning using sEMG signal that can be easily acquired by the portable device.
The EasiCSDeep comprehensively characterizes the high dimension sEMG signal with the low dimension features data, reorganize data into matrix format to support the computation of the deep learning, and developed the EasiDeep with six-channels to process the five types of feature data simultaneously which are able to automatically capture the dependency relationship and boost the performance of the model.
Compared with the previous study, the EasiCSDeep provides the state-of-the-art performance, outperforming the traditional learning algorithms.
The accuracy, sensitivity and specificity have been respectively improved by 6.2, 1.78 and 11.43 percentage points.
For individuals, the improvement in the accuracy decrease the misdiagnosis rate.
For the population, the improvement in sensitivity boosts the early detection of the more patient population. The improvement in specificity decreases the misdiagnosis rate of the healthy free of the CS. It is a very significant improvement which promotes the development of intelligent CS screening services in complex environments and boosts the study of the fine-grained classification models for the precision CS screening services.

In the future,
we will build a more comprehensive database of CS which involves a variety of the CS subtype and includes
multiple data types including images and medical records as well as expert knowledge and experience.
Based on the database above, we will
develop more intelligent and fine-grained classification system for CS screening to provide the convenient, low-cost and high-quality medical server, improving disease cure rate and reducing cost.
Meanwhile, we look forward to discovering more objective physiological indicators of cervical spondylosis development to further provide assistance for its diagnosis and treatment.

\end{document}